\documentclass[final]{cvpr}

\usepackage{times}
\usepackage{epsfig}
\usepackage{graphicx}
\graphicspath{{./}}
\usepackage{amsmath}
\usepackage{amssymb}

\usepackage{enumitem}
\usepackage{subfigure}
\usepackage{bm}
\usepackage{multirow}
\usepackage{multicol}
\usepackage{array}
\usepackage{booktabs}

\usepackage[pagebackref=true,breaklinks=true,colorlinks,bookmarks=false]{hyperref}

\def\cvprPaperID{7579} 
\def\confYear{CVPR 2021}
\begin{document}

\title{MetricOpt: Learning to Optimize Black-Box Evaluation Metrics}

\author{Chen Huang\qquad Shuangfei Zhai\qquad Pengsheng Guo\qquad Josh Susskind\\
Apple Inc.\\
{\tt\small \{chen-huang,szhai,pengsheng\_guo,jsusskind\}@apple.com}
}

\maketitle

\begin{abstract}

We study the problem of directly optimizing arbitrary non-differentiable task evaluation metrics such as misclassification rate and recall. Our method, named MetricOpt, operates in a black-box setting where the computational details of the target metric are unknown. We achieve this by learning a differentiable value function, which maps compact task-specific model parameters to metric observations. The learned value function is easily pluggable into existing optimizers like SGD and Adam, and is effective for rapidly finetuning a pre-trained model. This leads to consistent improvements since the value function provides effective metric supervision during finetuning, and helps to correct the potential bias of loss-only supervision. MetricOpt achieves state-of-the-art performance on a variety of metrics for (image) classification, image retrieval and object detection. Solid benefits are found over competing methods, which often involve complex loss design or adaptation. MetricOpt also generalizes well to new tasks and model architectures.

\end{abstract}

\section{Introduction}

In real-world vision applications, machine learning models are usually evaluated on a variety of complex evaluation metrics. For example, one may evaluate a classification model using Mis-Classification Rate (MCR), and evaluate a ranking model using recall. Many of these metrics are non-continuous, non-differentiable, or non-decomposable, which poses challenges for direct metric optimization due to the difficulty of obtaining an informative gradient (\eg,~it is zero almost everywhere for MCR). In other scenarios, the computational details may be unknown for a black box metric function. Hence its true gradient is simply inaccessible, which further increases the challenge of metric optimization.

As a common practice, people usually rely on a surrogate differentiable loss, which can be easily optimized with Stochastic Gradient Descent (SGD). One example is the widely used cross-entropy loss~\cite{Goodfellow-et-al-2016} for classification problems. While cross-entropy loss can be regarded as a smooth relaxation of MCR, this loss is not a good proxy for other metrics like recall. When the loss does not match the target metric, inferior performance can be obtained~\cite{ALA_2019}. State-of-the-art approaches follow two main paradigms to address this loss-metric mismatch issue. One is to introduce better metric-aligned surrogate losses,~\eg,~AUCPR loss~\cite{pmlr-v54-eban17a}. These hand-designed losses not only require tedious manual effort and white-box metric formulation, but also tend to be specific to a given metric. Another paradigm is to learn adaptive losses in a relaxed or interpolated surrogate space~\cite{abs-1905-10108,ALA_2019,jiang2020optimizing,liu2020unified}, which is inherently sub-optimal when compared to optimization in the original space.

\begin{figure}[!t]
\begin{center}
\centerline{\includegraphics[width=\columnwidth]{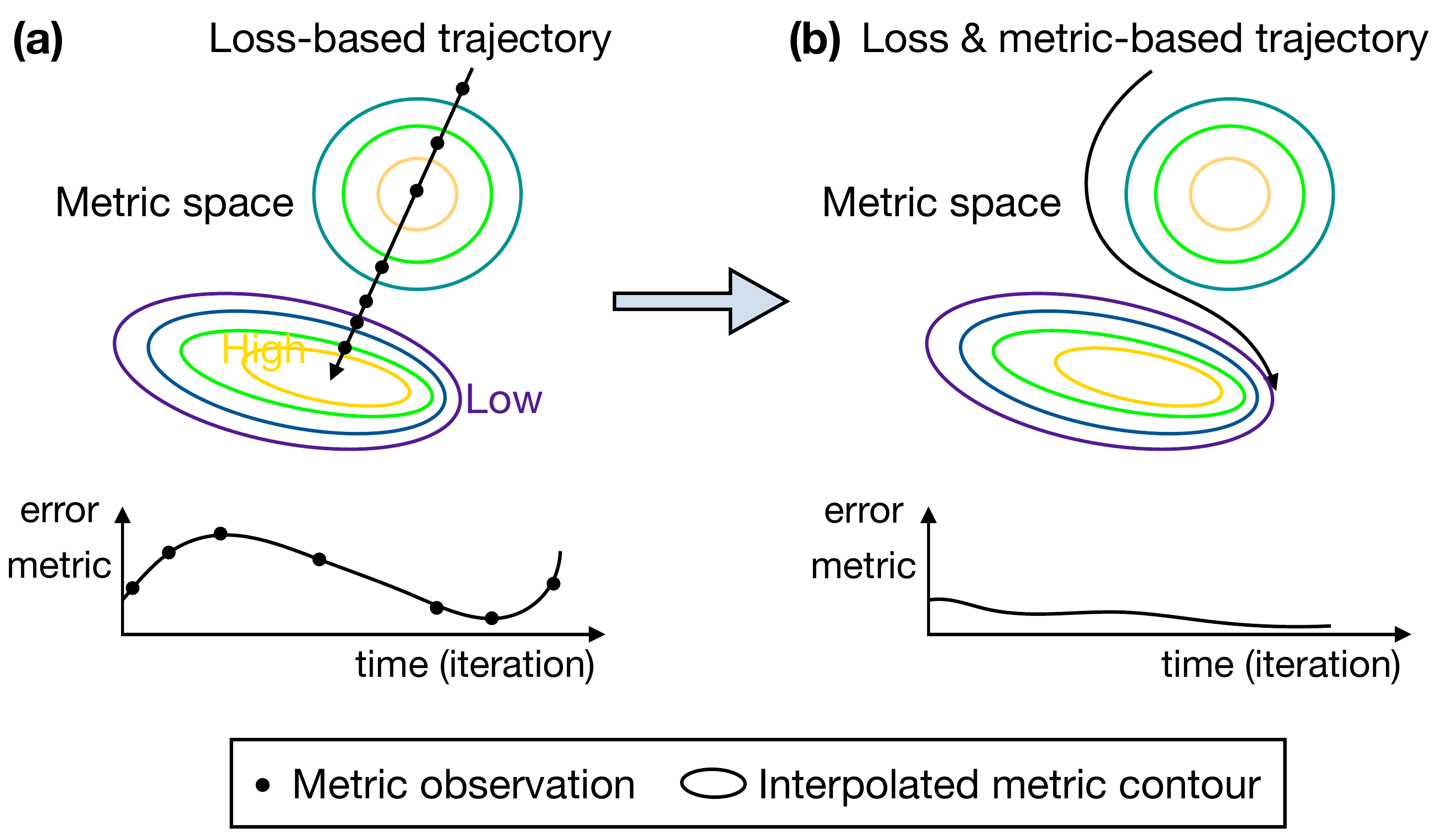}}
\caption{\textbf{Motivation} (a) Optimization with loss-only supervision may travel through several ``bumps'' in the metric space, and tends to converge to a suboptimal solution in terms of evaluation metric. We propose to collect a sparse set of (black-box) metrics along the optimization trajectory. Then we use the temporally interpolated metrics to meta-learn a differentiable value function, which can provide effective metric supervision to augment and improve any loss optimization process. (b) This way, we find continuous decrease in the error metric while keeping the loss from growing.}
\label{fig:Loss-metric-landscapes}
\end{center}
\vskip -0.4in
\end{figure}

In this paper, we propose to directly adapt the gradient-based optimization process to optimize \emph{black-box} metrics, without knowing any details about the metric function. To do so, we first meta-learn a differentiable value function to model the metric observations along optimization trajectories. We focus on the model finetuning setup which can provide meaningful metrics to learn our value function. Once learned, the value function can provide useful metric supervision, including approximate metric gradients, to augment surrogate loss gradients. As a result, we can use the value function to finetune a new model that has been pre-trained using any given surrogate loss. Fig.~\ref{fig:Loss-metric-landscapes} illustrates the high level idea. Intuitively, the value function is trained to produce meaningful adjustments to the optimization trajectory driven by loss only, leading to corrective directions on the metric landscape.

In practice, we parameterize the value function by a lightweight network for fast training and inference speeds. The input are a small set of \emph{adapter} parameters that modulate a pre-trained model. Such a compact parameterization has been shown to suffice for task specialization~\cite{zintgrafcavia}, and removes our need to finetune the entire network. We meta-learn our value function using an enhanced ordinal regression objective, which is uncertainty-aware to avoid overconfident metric estimates. We then show that it is straightforward to apply our value function to off-the-shelf optimizers like SGD and Adam~\cite{KingmaB15}, and also to a learned optimizer.

The resulting method \emph{MetricOpt} is shown to consistently improve different evaluation metrics across the tasks of (image) classification, image retrieval and object detection. MetricOpt not only outperforms prior methods based on strong surrogate losses (either hand-designed or adaptively learned), but also shares speed advantages as a fast finetuning method, often with no more than thousands of tuning steps. Furthermore, MetricOpt generalizes well to new tasks,~\eg,~from CIFAR-10~\cite{Krizhevsky09learningmultiple} to ImageNet~\cite{DenDon09Imagenet} classification. We summarize our contributions as follows:
\begin{itemize}[leftmargin=10pt]
\setlength{\itemsep}{0pt}
\setlength{\parsep}{0pt}
\setlength{\parskip}{0pt}
\item We introduce a differentiable value function to model a black-box evaluation metric.
\item We show the value function is easily pluggable to existing optimizers, resulting in a fast finetuning approach.
\item We show MetricOpt consistently improves over different surrogate losses without tedious loss engineering, achieves state-of-the-art performance for various tasks and metrics, and generalizes to out-of-distribution tasks and model architectures.
\end{itemize}

\section{Related Work}
\noindent
\textbf{Optimizing evaluation metrics} Due to the non-differentiable nature of most real-life evaluation metrics, a large body of surrogate loss functions have been proposed as smooth metric relaxations. Examples include AUCPR loss~\cite{pmlr-v54-eban17a}, pairwise AUCROC loss~\cite{Rakotomamonjy2004OptimizingAU}, Lov{\'a}sz-Softmax loss~\cite{berman2018lovasz} for IoU metric, and cost-sensitive classification for F-measure~\cite{NIPS2014_5508}.
To remove the manual effort to design these metric-approximating losses, there has been a recent push towards learning them instead~\cite{NIPS2018_7882,xu2018autoloss}. However, the loss learning is still based on metric relaxation schemes, whereas our focus is on a direct metric representation to be used for optimization.
Existing options for direct metric optimization are post-shift methods~\cite{NIPS2014_5454,NIPS2014_5504} that tune model threshold accordingly, and direct loss minimization methods~\cite{NIPS2010_4069,song2016training} that embed the true metric as a correction term for optimization. Their common limitation is that they require the evaluation metric to be available in closed-form.
Some recent works aim to optimize black-box metrics, from learning adaptive losses~\cite{abs-1905-10108,ALA_2019,jiang2020optimizing,liu2020unified} to learning adaptive example weights~\cite{pmlr-v97-zhao19b}. However, these methods can be sub-optimal since they work in a relaxed surrogate space or with predefined weighting schemes. By contrast, we learn to directly model black-box metrics which can in turn adapt the optimization process.

\noindent
\textbf{Computer vision metrics} are frequently defined as rank-based ones, such as recall for image retrieval, and Average Precision (AP) for object detection and feature matching,~\etc. Prior works approach the AP metric optimization via histogram binning approximations~\cite{Cakir_2019_CVPR,KHe18,He_2018_DOAP} and gradient approximation~\cite{HendersonF16,Mohapatra18,song2016training}. While these works mainly focus on AP relaxation, we differ in learning a fast and generic function approximation of the \textit{original} metric. Other techniques have been proposed to learn the ranking operation with a large LSTM~\cite{Engilberge_2019_CVPR}, or optimize an AP-loss using error-driven update~\cite{KChen19}, but both suffer from low efficiency. More recently, black-box differentiation~\cite{Rolinek_2020_CVPR} is used for efficient interpolation of the ranking function. More similar to our work is the deep embedding method~\cite{patel2020learning} where the model and embedding space are learned alternatively such that the Euclidean distance between the embedded prediction and groundtruth approximates the metric value. Our paper introduces a value function to regress metrics directly. The value function can be learned and deployed separately, runs fast and generalizes.

\noindent
\textbf{Meta learning} The goal of meta-learning is to infer a learning strategy from a distribution of similar tasks, which facilitates efficient learning of new tasks from the same distribution. Previous works have approached this problem by learning a good initialization~\cite{finn17a,abs-1803-02999,zintgrafcavia}, preconditioning matrix~\cite{abs-1909-00025} or mathematical update equation~\cite{pmlr-v70-bello17a}. We use the efficient Reptile algorithm~\cite{abs-1803-02999} to meta-learn the initialization of our value function in an online fashion.

\noindent
\textbf{Value function approximation} Many Reinforcement Learning (RL) methods require value function approximation to estimate an action-value for fast decision making. Popular off-policy actor-critic methods~\cite{fujimoto2018addressing,haarnoja18b} are such examples. Their critic uses a differentiable function approximator to provide a loss, which guides the actor to update its action policy. The notion of ``value function'' in RL is related to the idea of learning a parametric loss as a metric approximator in the optimization field~\cite{ALA_2019,abs-1905-10108}, in order to provide some proxy gradients. Here we directly learn a differentiable mapping to evaluation metrics, bypassing the tedious step of loss function engineering.

\begin{figure*}[!t]
\begin{center}
\centerline{\includegraphics[width=1.0\linewidth]{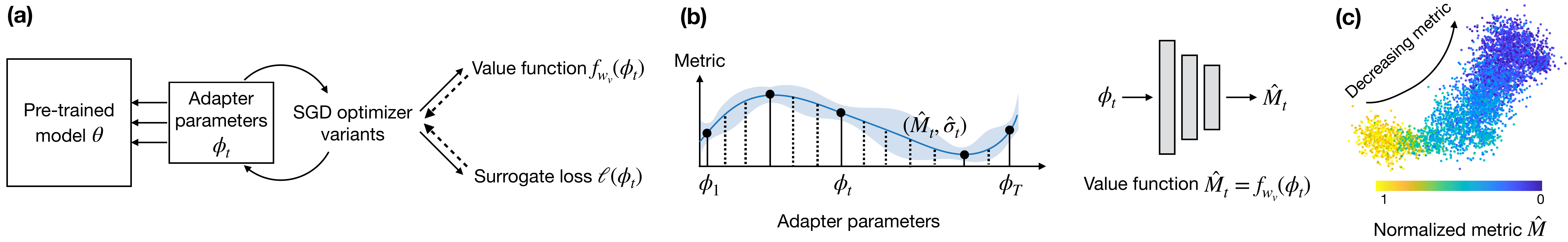}}
\caption{(a) Alongside finetuning the adapter parameters $\phi$ (\eg,~FiLM conditioning parameters~\cite{perez2018film}) of a pre-trained model $\theta$, we learn a value function $f_{w_v}$ that maps $\phi$ to the non-differentiable metric $\mathcal{M}$. The learned value function is applicable to existing optimizers (\eg,~SGD and Adam) in a plug-and-play manner to augment loss-based optimization. (b) We parameterize $f_{w_v}$ by a lightweight network, which is meta-trained from different finetuning trajectories (solid stems -- sparse metric observations, dotted lines -- interpolated metrics $\hat{\mathcal{M}}_t$ and variances $\hat{\sigma}_t$). (c) t-SNE~\cite{vanDerMaaten2008} visualization of 2D feature embeddings of $f_{w_v}$, where the ordinal ranks of metrics are well preserved.}
\label{fig:value_func}
\end{center}
\vskip -0.4in
\end{figure*}

\section{Methodology}

We consider optimizing the evaluation metric $\mathcal{M}$ of a neural network model in a finetuning setup, where the model weights $\theta$ have been pre-trained using a user-specified surrogate loss. Instead of finetuning the high-dimensional $\theta$ which is costly, we optimize the conditioning parameters $\phi \in \Phi$ of a small adapter module that modulates $\theta$ (detailed later). Assume the surrogate loss is given as $\ell(\theta,\phi)$ for the tuple $(\theta,\phi)$. For brevity, we will, moving forward, use $\ell(\phi)$ in place of $\ell(\theta,\phi)$ since $\theta$ is fixed. Let $\mathcal{M}(\phi)$ be the metric subject to minimization. We also assume $\ell(\phi)$ and $\mathcal{M}(\phi)$ are normalized into the range of $[0,1]$ for scale-independent modeling across tasks.

Generally, the metric function of $\mathcal{M}$ is either non-differentiable (true gradient of the metric is often not meaningful), or simply unknown (black box setting with no access to the gradient). In both cases, we are faced with the great challenge of direct metric optimization. Here, we assume the only operation available to us is the ability to query metrics from a meaningful set of examples, without knowing the computational details of $\mathcal{M}$. Evaluation can be conducted on a held-out validation set $D_{val}$ or on the training set $D_{train}$. We consider optimizing the metric in the form:
\begin{equation}
\phi^* = \arg \min_{\phi\in\Phi} \lambda \mathcal{M}(\phi) + \ell(\phi),
\label{eq1}
\end{equation}
where $\lambda$ is a weighting parameter of $\mathcal{M}(\phi)$, and loss $\ell(\phi)$ acts as an auxiliary signal. Without metric supervision, Eq.~\eqref{eq1} is reduced to the standard loss optimization problem, which often leads to the loss-metric mismatch and hence suboptimal performance in evaluation metrics. On the other hand, without the surrogate loss, the supervision from metric alone is often sparse and limiting. We will show in the paper that the metric supervision always boosts the loss-based performance without any bells and whistles.

\subsection{Meta learning differentiable value function}
\label{sec:value_function}

In Eq.~\eqref{eq1}, we aim at efficient optimization of the metric term $\mathcal{M}(\phi)$ via gradient-based methods. The core component of our approach is the value function $f_{w_v}: \Phi \rightarrow \mathbb{R}$, a differentiable mapping from input $\phi\in\Phi$ to metric $\mathcal{M}\in \mathbb{R}$. We parameterize $f$ by a deep neural network with weights $w_v$. It has the benefits of being able to output both metric estimate for any $\phi$, as well as partial derivatives $\partial f/\partial \phi$~\wrt~parameters $\phi$ to augment the loss gradients $\partial \ell/\partial \phi$. In other words, the value function is a generic function approximation of $\mathcal{M}$, which can offer useful supervision for the metric and gradient-based optimizablility. As a result, finetuning can be accomplished using both the loss function and value function, without a significant computational overhead, see Fig.~\ref{fig:value_func}(a).


In practice, we train the value function from different optimization tasks $\mathcal{T} \sim p(\mathcal{T})$, where each task finetunes randomly initialized $\phi$ with random mini-batches (two sources of task variability). The finetuning process ensures the value function can be learned with meaningful metrics around converged $\theta$, rather than with noisy metrics collected by a ``from-scratch'' model. This simplifies value function learning and gives rise to a fast finetuning process. Below we address two main challenges for value function learning.

\textbf{Adapter parameters} The high dimensionality of input parameters $\theta$ (often millions) renders value function learning parameter-inefficient. To this end, we have tried using a subset of $\theta$ in early experiments,~\eg,~only the layer biases or last layer. However, we found these approaches still not parameter-efficient enough, and they cannot strike a good performance-efficiency tradeoff.

In this paper, we turn to using the \emph{adapter} modules that offer an equally simple but better performing solution. Existing adapters like conditional BatchNorm~\cite{NIPS2017_7237} and FiLM layers~\cite{perez2018film} introduce a few conditioning parameters $\phi \in \mathbb{R}^d$ to modulate the fixed $\theta$. They have been shown to suffice for task specialization, with comparable performance to full model fine-tuning. We follow~\cite{zintgrafcavia} to use dynamic biases at the first layer of fully connected networks, and use FiLM parameters to modulate the feature maps of convolutional networks (visualizations in supplementary materials). These two types of $\phi$ have small dimension $d$ on the order of tens to hundreds, and are found effective in our experiments. We are open to other choices.

\begin{figure*}[!t]
\begin{center}
\centerline{\includegraphics[width=0.7\linewidth]{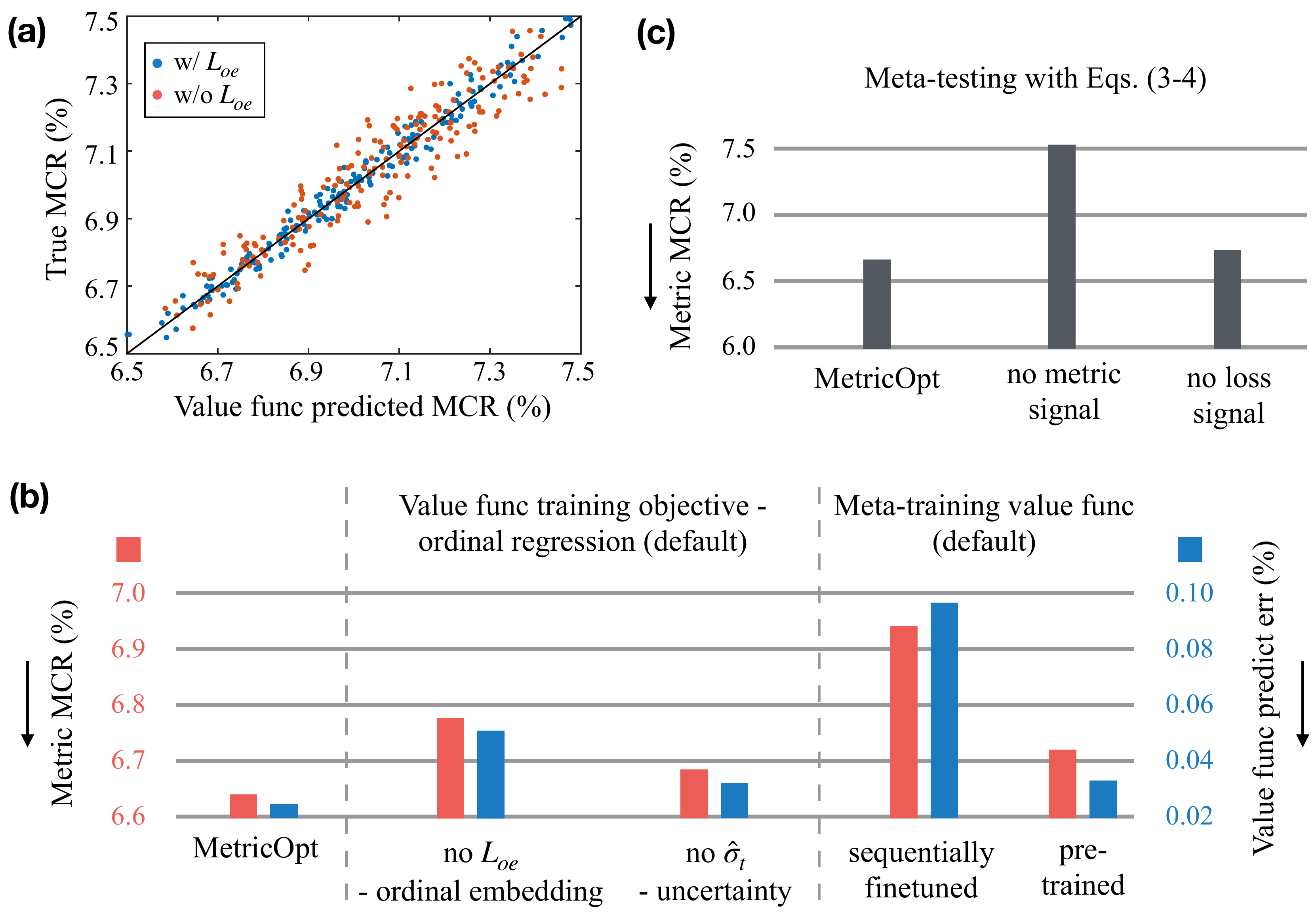}}
\caption{Optimizing Mis-Classification Rate (MCR) on CIFAR10. (a) Value function trained with and without the ordinal embedding constraint $L_{oe}$. The latter has more scattered predictions and higher prediction error (0.052 vs. 0.023). (b) Value function prediction error highly correlates with the final optimized metric, thus can serve as a performance indicator to diagnose~\eg,~meta-training alternatives for the value function. (c) Comparing different meta-testing schemes.}
\label{fig:ECE_analysis}
\end{center}
\vskip -0.4in
\end{figure*}

\textbf{Metric interpolation} Another challenge for value function training is the lack of training signals. Usually we can only collect $K \ll T$ metrics $\{\mathcal{M}_k=\mathcal{M}(\phi_k)\}_{k=1}^K$ during $T$ finetuning steps, where fast metric evaluation may not be an option. For instance, metrics like recall require evaluation on sizable data to be statistically significant, and the large evaluation cost makes frequent evaluation impractical. For this reason, we simply use the $K$ sparse metric observations, with each $\mathcal{M}_k$ evaluated at user-specified time $k \in [1,\dots,T]$. We further interpolate metrics over time to enrich training signals for our value function. This incurs much lower cost than using heavier metric evaluations, but has similar performance empirically. Specifically, we use a standard Gaussian process with RBF kernel~\cite{GP2006}, and interpolate metrics into a dense time series $\{(\hat{\mathcal{M}}_t,\hat{\sigma}_t)\}_{t=1}^T$ with mean $\hat{\mathcal{M}}_t$ and variance $\hat{\sigma}_t$ at time $t$, see Fig.~\ref{fig:value_func}(b). Compared to linear interpolation like line fitting, the Gaussian process can derive useful uncertainties for the interpolated metrics, which is important to avoid overconfident supervision for value function learning.

\textbf{Value function parameterization} Given sequence $\{\phi_t\}_{t=1}^T$ and the corresponding ``labels'' $\{(\hat{\mathcal{M}}_t,\hat{\sigma}_t)\}_{t=1}^T$, we can now train the mapping $\hat{\mathcal{M}}_t=f_{w_{v}}(\phi_t)$ as our value function. We parametrize $f_{w_v}$ by an MLP (Multi-Layer Perceptron) network with architecture d--64--32--32--16--1. BatchNorm and ReLU activation are used for all layers. The network is lightweight for fast metric regression and gradient-based optimization, but can further benefit from improved design choices for particular problems or metrics.

\textbf{Value function training objective} consists of a regression term $L_{regress}$ and an ordinal embedding term $L_{oe}$:
\begin{eqnarray}  
\!\!\!\! L_v(w_v) \!\!\!\!\!\!\!\!\!\! &&= \gamma L_{regress} + L_{oe}, \\ \nonumber
\!\!\!\!  L_{regress} \!\!\!\!\!\!\!\!\!\! &&= \frac{1}{\sum_{t=1}^T 1/\hat{\sigma}_t} \sum_{t=1}^T \frac {\|f_{w_v}(\phi_t)-\hat{\mathcal{M}}_t\|_2}{\hat{\sigma}_t}, \\ \nonumber
\!\!\!\!  \mathcal{L}_{oe} \!\!\!\!\!\!\!\!\!\! &&= \frac{1}{T} \sum_{t=1}^T \log \left( 1+ \exp(-(D_{t,t_n} -D_{t,t_p} )) \right), \nonumber
\label{eq2}
\end{eqnarray}
where $\gamma$ is a weighting parameter, $D_{t,t'}=\|g_{w_v}(\phi_t)-g_{w_v}(\phi_{t'}) \|_2$ is the Euclidean distance between the feature embeddings $g_{w_v}(\cdot)$ of our value function (penultimate layer), and $(t,t_p,t_n)$ is the sampled triplet at time $t$ for embedding learning.

Note $L_{regress}$ is aware of the uncertainty $\hat{\sigma}_t$ that helps to re-weight metric regression errors. While the $L_{oe}$ term imposes ordinal constraints for the metric in feature space. Concretely, we sample triplets $(t,t_p,t_n)$, with anchor $t \in [1,\dots,T]$, its positive sample $t_p \in \mathcal{P}_t$ and negative sample $t_n \in \mathcal{N}_t$ for embedding learning. To do so, we leverage Fisher's ratio $r_{t,t'}=(\hat{\mathcal{M}}_{t}-\hat{\mathcal{M}}_{t'})^2/ (\hat{\sigma}_{t}^2+\hat{\sigma}_{t'}^2)$ that measures the discrimination between two metric variables. The Fisher's ratio offers us great convenience to construct the positive set $\mathcal{P}_t = \{t' | r_{t,t'}<2\}$ and negative set $\mathcal{N}_t = \{t' | r_{t,t'}\ge 2\}$ for every time step $t$. Then we perform hard mining within $\mathcal{P}_t$ and $\mathcal{N}_t$ to sample the required $t_p$ and $t_n$, in a similar way to~\cite{wang2019multi}. In general, $L_{oe}$ provides effective regularization in feature space (see Fig.~\ref{fig:value_func}(c)) and eases our value function training. Fig.~\ref{fig:ECE_analysis}(a-b) shows improved performance by $L_{oe}$. Fig.~\ref{fig:ECE_analysis}(b) also validates the need to encode metric uncertainties $\hat{\sigma}_t$ for value function learning.


\textbf{Meta-training value function} For meta-training $f_{w_v}$, we have access to different finetuning trajectories of tasks $\mathcal{T} \sim p(\mathcal{T})$ and their densely interpolated metrics. Task variability comes from random initialization $\phi_0$ and mini-batches. Then using the given training data and objective (Eq.~(2)), how do we train $f_{w_v}$ effectively and efficiently? We follow the first-order Reptile algorithm~\cite{abs-1803-02999} to meta-update $w_v$ in an online fashion, see Algorithm~1 (meta-training stage). Such meta-training avoids catastrophic forgetting with sequential task finetuning by knowledge distillation into the initialization of $w_v$. Another competing scheme is to pre-train the value function from all the task trajectories stored offline, which however requires constructing a large dataset. Fig.~\ref{fig:ECE_analysis}(b) compares performance for the two schemes. 

\subsection{Meta-testing with learned value function}
\label{sec:SGD}

Meta-testing involves optimizing the target metric for a new task $\mathcal{T}$,~\ie,~finetuning a new pre-trained model $\theta$ or with a different adapter initialization $\phi_0$. The key here is to use our value function to augment a given surrogate loss with metric information during finetuning. Recall that value function can provide direct metric supervision or explicit gradients about metric. Hence it can be readily applied to off-the-shelf optimizers in a plug-and-play manner.

Inspired by the Guided ES approach~\cite{maheswaranathan19a}, we combine our value function with SGD/Adam optimizers in a random search framework. We call the resulting method Metric Optimizer (MetricOpt). The high-level idea is to perform value function-based random search around surrogate loss gradients, in order to estimate a new descent direction favoring metric. Specifically, we keep track of a subspace defined by the past $k$ loss gradients from SGD or Adam. Then the orthonormal basis $U\in \mathbb{R}^{d\times k}$ of the subspace is derived. We define the search covariance as:
\begin{equation}
\Sigma = \frac{1}{2d}I+\frac{1}{2k}UU^T.
\label{eq3}
\end{equation}
By sampling $P$ perturbations $\{\delta_i\}_{i=1}^P \sim \mathcal{N}(0,s^2\Sigma)$ with variance $s^2$ within the subspace, we estimate the descent direction $u_t$ using Evolutionary Strategies (ES) informed by our learned value function $f_{\tilde{w}_v}$:
\begin{equation}
u_t= \frac{1}{s^2P} \sum_{i=1}^P \delta_i \left[ f_{\tilde{w}_v}(\phi_t+\delta_i)- f_{\tilde{w}_v}(\phi_t-\delta_i) \right].
\label{eq4}
\end{equation}



\begin{figure}[!t]
\begin{center}
\centerline{\includegraphics[width=0.95\columnwidth]{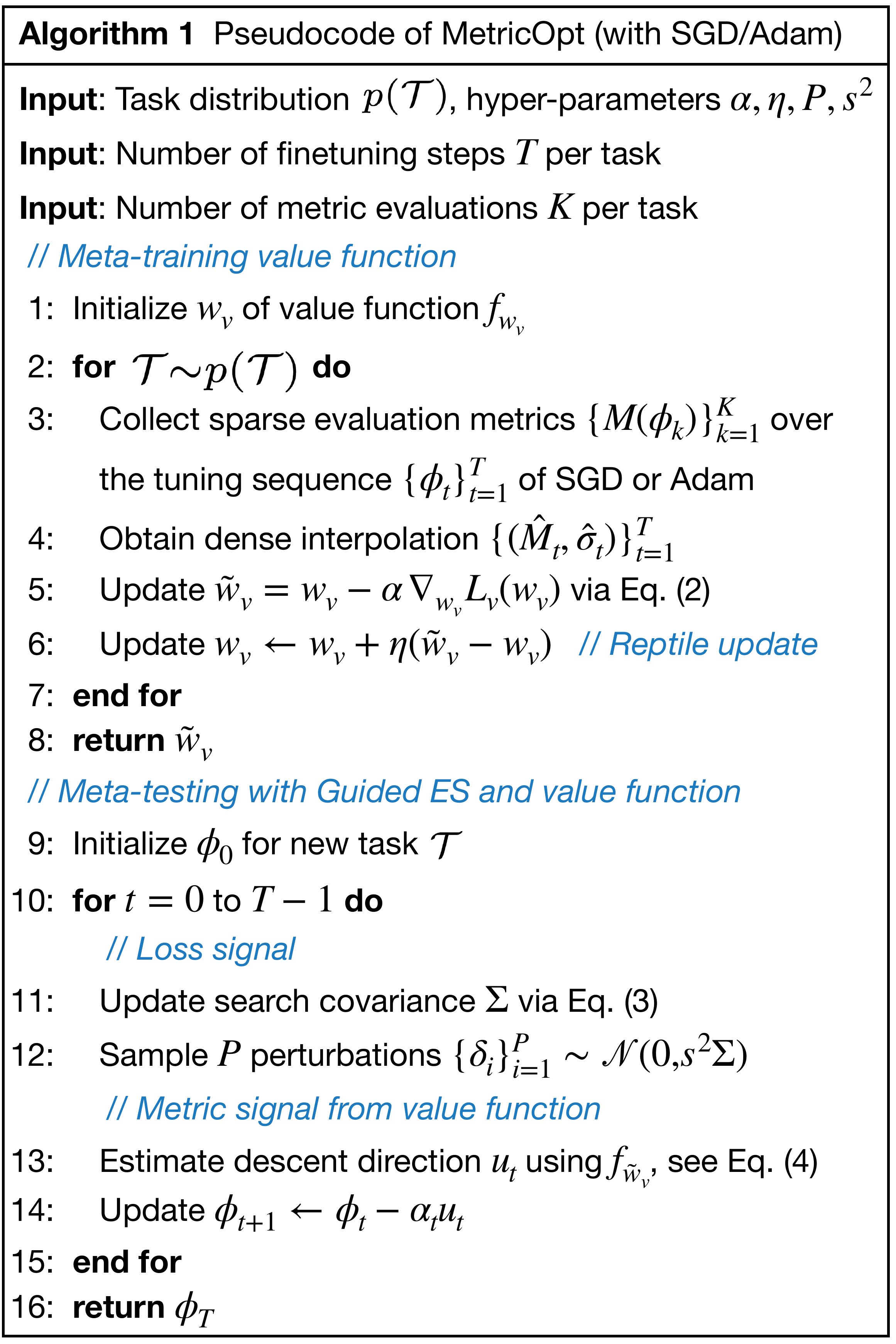}}
\label{fig:algorithm}
\end{center}
\vskip -0.4in
\end{figure}

\textbf{Remarks} The above meta-testing method provides one simple way to combine the loss and metric signals for finetuning. As evidenced in Fig.~\ref{fig:ECE_analysis}(c), the metric signal plays an important role. Without it, we see a large performance drop as expected. Whereas using metric alone (no auxiliary loss) barely hurts performance. The downside is the reduced efficiency -- tuning with metric only will incur a larger cost for random search (over 5$\times$ slower in the CIFAR10 case). Therefore we always treat our value function as a metric improver for surrogate losses, including those well-designed and learned losses. Also, the value function can be easily applied to other gradient-based optimization methods. Supplementary materials give one example of combining value function with a learned optimizer that achieves state-of-the-art performance.

\begin{table*}[!t]

\begin{minipage}{.5\textwidth}
\caption{Mis-Classification Rate (MCR) on CIFAR-10. The average and standard deviation are from 10 runs. Note MetricOpt (SGD) is our default approach, while MetricOpt (learned) involves joint learning of value function and optimizer (see text for details).}
\label{tb:cifar10_MCR}
\begin{center}
\begin{small}
\begin{tabular}{lc}
\hline
Method & MCR (\%) $\downarrow$ \\
\hline
Cross-entropy loss & 7.51 \\
Large-margin softmax~\cite{liu2016large} & 7.01 \\
L2T-DLF~\cite{NIPS2018_7882} & 6.95 \\ \hline
Example weighting (MOEW)~\cite{pmlr-v97-zhao19b} & 6.71$\pm$0.08 \\ 
Adaptive Loss Alignment (ALA)~\cite{ALA_2019} & 6.79$\pm$0.07 \\ 
Black-box differentiation~\cite{Rolinek_2020_CVPR} & 6.84$\pm$0.05 \\ \hline
Cross-entropy finetune + dense metric eval & 7.19$\pm$0.16 \\ \hline
MetricOpt (SGD) + cross-entropy & 6.63$\pm$0.05 \\ 
MetricOpt (learned) + cross-entropy & 6.58$\pm$0.06 \\
MetricOpt (SGD) + large-margin softmax~\cite{liu2016large} & 6.56$\pm$0.06 \\ 
MetricOpt (learned) + large-margin softmax~\cite{liu2016large} & \textbf{6.47$\pm$0.07} \\
\hline
\end{tabular}
\end{small}
\end{center}
\vskip -0.2in
\end{minipage}
\hspace{0.1in}
\begin{minipage}{.5\textwidth}
\caption{Area Under the Precision Recall Curve (AUCPR) on CIFAR-10. The average and standard deviation are from 10 runs. MetricOpt (SGD) is our default approach, while MetricOpt (learned) involves joint learning of value function and optimizer (see text for details).}
\label{tb:cifar10_AUCPR}
\begin{center}
\begin{small}
\begin{tabular}{lc}
\hline
Method & AUCPR (\%) $\uparrow$ \\
\hline
Cross-entropy loss & 84.6 \\
Pairwise AUCROC loss~\cite{Rakotomamonjy2004OptimizingAU} & 94.2 \\
AUCPR loss~\cite{pmlr-v54-eban17a} & 94.2 \\ \hline
Example weighting (MOEW)~\cite{pmlr-v97-zhao19b} & 94.6$\pm$0.11 \\ 
Adaptive Loss Alignment (ALA)~\cite{ALA_2019} & 94.9$\pm$0.14 \\
Black-box differentiation~\cite{Rolinek_2020_CVPR} & 94.4$\pm$0.07 \\ \hline
Cross-entropy finetune + dense metric eval & 90.8$\pm$0.18 \\ \hline
MetricOpt (SGD) + cross-entropy & 95.7$\pm$0.12 \\ 
MetricOpt (learned) + cross-entropy & 96.2$\pm$0.09 \\
MetricOpt (SGD) + AUCPR loss~\cite{pmlr-v54-eban17a} & 96.6$\pm$0.10 \\ 
MetricOpt (learned) + AUCPR loss~\cite{pmlr-v54-eban17a} & \textbf{97.2$\pm$0.08} \\
\hline
\end{tabular}
\end{small}
\end{center}
\vskip -0.2in
\end{minipage}
\vskip -0.2in
\end{table*}

\subsection{Computational complexity}
Algorithm 1 shows meta-training of our value function over three main steps: obtain the finetuning trajectory for task $\mathcal{T}$ with task-dependent tuning steps $T$, metric interpolation for task $\mathcal{T}$, and meta-update of the value function. The task finetuning is very efficient since in practice we find it only involves hundreds of tuning steps over the compact adapter parameters $\phi$. The overhead of metric interpolation is negligible. Overall, our value function can be trained for 0.5k to 2k meta-iterations (tasks), which requires less than half a GPU day for all experiments reported. For meta-testing, finetuning with Guided ES and value function is only 22\% to 34\% slower than standard loss optimization. The overhead is reasonable given the few finetuning steps.

\section{Experiments}

We aim to answer the following questions in our experiments: 1) Does the value function consistently improve over surrogate losses on problems with only black-box access to the metric? 2) How does MetricOpt compare to state-of-the-art methods for metric optimization, and why the difference? 3) Can MetricOpt generalize to new tasks and model architectures? To answer these questions, we consider optimizing various evaluation metrics on a diverse set of problems, including (image) classification, image retrieval and object detection.

\textbf{Hyper-parameters} In all experiments we set $\gamma=10$ to make the two terms in Eq.~(2) roughly equally weighted. Results are insensitive to $\gamma$ in a wide range. The number of finetuning steps $T$ is task-dependent. For meta-testing with Guided ES~\cite{maheswaranathan19a}, we use the same parameters---search within a subspace of past $k=3$ loss gradients (Eq.~(3)), and sample $P=3$ perturbations with variance $s^2=0.01$ (Eq.~(4)). In Algorithm 1, the inner learning rate $\alpha=0.005$, and meta learning rate $\eta=1$ with a linear decay to 0.

\begin{figure}[!t]
\begin{center}
\centerline{\includegraphics[width=1.0\columnwidth]{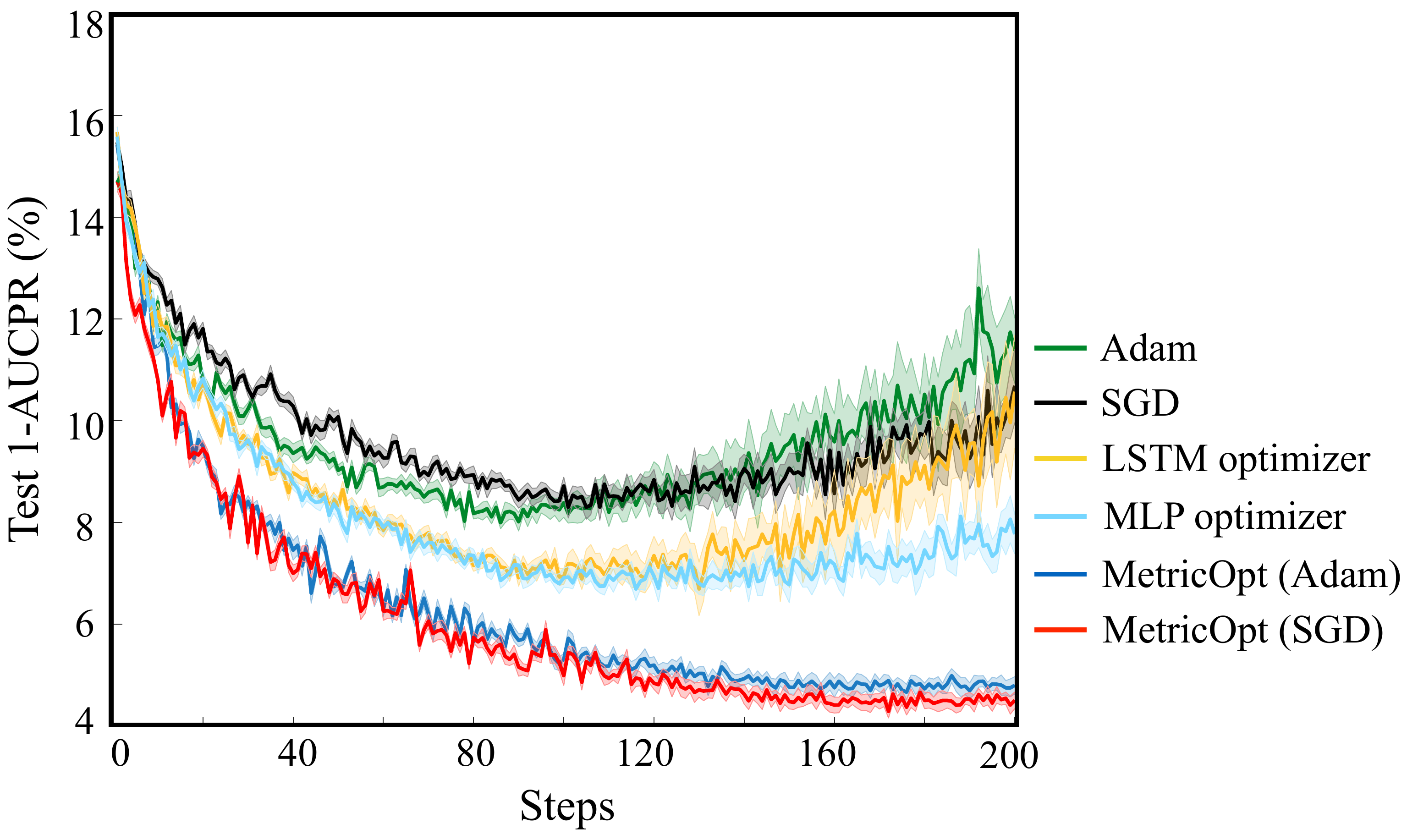}}
\caption{10-run results of optimizing AUCPR on CIFAR-10. The metric undergoes a $1-x$ conversion, so the lower the better. Our MetricOpt, when combined with either SGD or Adam, outperforms both the SGD variants and learned optimizers on surrogate cross-entropy loss.}
\label{fig:CIFAR10_res}
\end{center}
\vskip -0.4in
\end{figure}

\subsection{Image classification}
We optimize the Mis-Classification Rate (MCR) and Area Under the Precision Recall Curve (AUCPR) on CIFAR-10 dataset~\cite{Krizhevsky09learningmultiple}. ResNet-32~\cite{He2016DeepRL} and the network in~\cite{pmlr-v54-eban17a} are used for MCR and AUCPR respectively, for fair comparisons with methods that use the same networks. We use the train/validation/test split of sizes 45k/5k/10k for both MetricOpt and those methods~\cite{NIPS2016_6461,ALA_2019,metz19a,pmlr-v97-zhao19b} whose training also requires querying the validation statistics (loss or metric). MetricOpt is trained to optimize $d=128$ conditioning parameters $\phi$ of the FiLM layers for considered ConvNets. During meta-training, we finetune $T=200$ steps and randomly collect $K=5\%T$ evaluation metrics on validation data $D_{val}$ along the finetuning trajectory. We report 10-run results of meta-testing on $D_{test}$.

First off, Fig.~\ref{fig:CIFAR10_res} uses the AUCPR case to compare our MetricOpt against existing popular optimizers. AUCPR is a non-differentiable metric and hard to optimize directly. Hence existing optimizers often choose the widely used cross-entropy loss as a surrogate, and we will discuss about advanced surrogates later. For best performance, we tune the learning rates of hand-crafted optimizers SGD and Adam~\cite{KingmaB15}, and tune the meta learning rates of the learned LSTM optimizer~\cite{NIPS2016_6461} and MLP optimizer~\cite{metz19a} by a grid search over the range of $[10^{-4}, 1]$. We can see from the figure that existing loss optimizers suffer from the absence of metric supervision and do not optimize the metric well. By contrast, MetricOpt helps SGD/Adam to converge to better metrics by the guidance of value function.


\begin{table*}[t]
\caption{Binary classification on A9A and CoverType (Cov) datasets. We compare MetricOpt (with cross-entropy loss) against learned Surrogate Loss (SL) and various metric-specific losses. Specifically, we compare against CE (Cross-Entropy) and SL for MCR (Mis-Classification Rate), LO (Lov{\'a}sz-Softmax loss) and SL for JAC (Jaccard Index), and CS (Cost-Sensitive) and SL for F-measure.}
\label{tb:UCI_datasets}
\begin{center}
\begin{small}
\begin{tabular}{cccccccccc}
\hline
\multirow{2}{*}{Data} & \multicolumn{3}{c}{MCR $\downarrow$} & \multicolumn{3}{c}{JAC $\uparrow$} & \multicolumn{3}{c}{F-measure $\uparrow$}\\
\cmidrule(lr){2-4}
\cmidrule(lr){5-7}
\cmidrule(lr){8-10}
 & CE & SL~\cite{abs-1905-10108} & MetricOpt & LO~\cite{berman2018lovasz} & SL~\cite{abs-1905-10108} & MetricOpt & CS~\cite{NIPS2014_5508} & SL~\cite{abs-1905-10108} & MetricOpt\\
\hline
A9A& 0.1520 & 0.1502 & \textbf{0.1369} & 0.8461 & 0.8488 & \textbf{0.8691} & 0.6823 & 0.6866 & \textbf{0.6934} \\
Cov& 0.2149 & 0.2198 & \textbf{0.2057} & 0.7406 & \textbf{0.7808} & 0.7758 & 0.7695 & 0.7898 & \textbf{0.8013} \\
\hline
\end{tabular}
\end{small}
\end{center}
\vskip -0.1in
\end{table*}

\begin{table*}[t]
\caption{Top-1 and Top-5 classification accuracies (\%) on ImageNet. All methods use the same NASNet-A network.}
\vskip 0.1in
\label{tb:transfer_cls}
\begin{center}
\begin{small}
\begin{tabular}{lcc}
\hline
Method & Top-1 & Top-5 \\
\hline
RMSProp + cross-entropy & 73.5 & 91.5 \\ 
PowerSign-cd~\cite{pmlr-v70-bello17a} (CIFAR-10 transfer) & 73.9 & 91.9  \\
ALA~\cite{ALA_2019} (CIFAR-10 transfer) & 74.3 & 92.1 
\\\hline
MetricOpt (transfer from ImageNet training with ResNet-18) & 74.2 & 92.3 \\
MetricOpt (transfer from CIFAR-10 training with NASNet-A) & 74.4 & 92.2 \\
MetricOpt (ImageNet training with NASNet-A) & \textbf{74.9} & \textbf{92.7} \\ \hline
MetricOpt + 2nd run (finetune $\theta$ and $\phi$) & \textbf{75.1} & \textbf{93.0} \\ \hline
\end{tabular}
\end{small}
\end{center}
\vskip -0.2in
\end{table*}

In the following, \textbf{we choose MetricOpt (SGD) as our default approach}. Tables~\ref{tb:cifar10_MCR} and~\ref{tb:cifar10_AUCPR} compare MetricOpt with advanced loss functions and state-of-the-art black-box metric optimization methods. We have several observations:
\begin{itemize}[leftmargin=10pt]
\setlength{\itemsep}{0pt}
\setlength{\parsep}{0pt}
\setlength{\parskip}{0pt}
\item Surrogate loss functions (top cell) are suboptimal, although they take lots of time for manual design or online learning (\eg,~L2T-DLF).
\item MOEW and ALA learn dynamic weighting schemes for data and loss respectively, both adapted to black-box metric observations. But MOEW is limited by using a predefined weighting scheme, and ALA still works in a relaxed surrogate space. In~\cite{Rolinek_2020_CVPR}, gradient interpolation is performed via black-box differentiation. Our MetricOpt is more competitive by directly adapting the optimization process based on explicit metric modeling.
\item With MetricOpt, our value function consistently improves over surrogate losses, including cross-entropy loss and advanced losses (large-margin softmax or AUCPR loss). The gains over advanced losses are smaller because they are often better aligned with metrics already. However, designing advanced losses requires tremendous human efforts. Hence in the following experiments, we mainly combine our value function with the default loss function for each problem, which waives the need for tedious loss engineering.
\item A learned optimizer from the loss and value functions finds extra boosts. Supplementary materials introduce the learning mechanism and more results.
\item We further compare with a simple finetuning baseline: it finetunes adapter parameters just like MetricOpt, but uses cross-entropy loss only and evaluates metrics at every step with the best picked. We see the resulting performance is far from that of MetricOpt. This suggests MetricOpt can find a truly different solution off loss-based trajectories. The solution has better test metric without requiring excessive metric evaluations along finetuning.
\end{itemize}


\subsection{Binary classification on non-image data}

Experiments are conducted on the A9A and CoverType datasets from the UCI Machine Learning Repository~\cite{Dua2017}. We randomly split data into 70\%/10\%/20\% as the train/validation/test sets for both datasets. All methods use class-balanced mini-batch sampling, and train a Leaky ReLU-activation MLP with 100-30-10-1 neurons. BatchNorm and dropout are applied at each layer. Our \mbox{MetricOpt} finetunes $d=16$ dynamic biases added at the input layer for $T=50$ steps. We still use cross-entropy loss as the baseline loss for model pre-training and \mbox{MetricOpt}-based finetuing, arriving at MetricOpt (SGD) + cross-entropy again.

Table~\ref{tb:UCI_datasets} compares our MetricOpt with competing methods for 3 evaluation metrics---MCR, JAC and F-measure. Specifically, we compare with hand-designed metric-specific losses CE, LO~\cite{berman2018lovasz} and CS~\cite{NIPS2014_5508}. We also compare with the learned Surrogate Loss (SL)~\cite{abs-1905-10108}, a representative of recent adaptively learned surrogate losses~\cite{abs-1905-10108,ALA_2019,jiang2020optimizing,liu2020unified}. The table generally confirms our advantages over these surrogate loss methods. This demonstrates the versatility of MetricOpt on black-box metrics.

\subsection{Generalization test on ImageNet}

This section compares different methods on the large-scale ImageNet dataset~\cite{DenDon09Imagenet}. All methods use the same NASNet-A network architecture~\cite{Zoph_2018_CVPR} unless otherwise noted. Evaluation metrics are the top 1 and top 5 classification accuracies. Table~\ref{tb:transfer_cls}, top cell, outlines the full model training baselines, while the middle cell lists our MetricOpt-based finetuning variants. All MetricOpt variants finetune $d=256$ conditioning parameters of FiLM layers for $T=500$ steps. They aim at augmenting the cross-entropy loss to improve the MCR metric. Note for fair comparison, the number of model pre-training plus finetuning steps of MetricOpt is the same as the total number of training steps for full training methods.

We can see from Table~\ref{tb:transfer_cls} that the RMSProp optimizer achieves suboptimal results since it merely optimizes the cross-entropy loss. PowerSign-cd~\cite{pmlr-v70-bello17a} and ALA~\cite{ALA_2019} transfer their update policy and loss controlling policy learned from CIFAR-10 to ImageNet, both showing strong results of out-of-distribution training. Our MetricOpt considers direct metric optimization during finetuning. In doing so, we obtain even better results than the full training baselines, although we only finetune a small number of parameters. We further study the transferability of MetricOpt by reusing our value function in two settings: learned with a smaller base network (ResNet-18) on the same ImageNet data, and learned with the same NASNet-A network on a smaller dataset (CIFAR-10). MetricOpt achieves better generalization results for a larger problem or model architecture, demonstrating the flexibility of the approach.

Given the good finetuning performance of MetricOpt, the bottom cell of Table~\ref{tb:transfer_cls} studies the following question: what if we also finetune the pre-trained network $\theta$ instead of keeping it fixed and only finetuning the modulating parameters $\phi$? To answer the question, we build on the finetuned $\phi$ in the first run, and continue to fine-tune $\theta$ (using cross-entropy loss) followed by a quick adjustment of $\phi$ via MetricOpt for a second run. This variant leads to marginal gains, but has much lower time efficiency than just tuning $\phi$ since finetuning the high-dimensional $\theta$ is very expensive. Moreover, we found diminishing returns when we alternate between finetuning $\theta$ and $\phi$ for a couple of more rounds.


\subsection{Image retrieval}

The image retrieval task is a perfect testbed for optimizing rank-based metrics like recall. We conduct experiments on the Stanford Online Products dataset~\cite{SongXJS16} which contains 120,053 images of 22,634 classes. We follow the standard train/test splits and image data preparation procedure in~\cite{SongXJS16}. A pretrained ResNet-50~\cite{He2016DeepRL} is used with a fully connected embedding layer. The embedding has a fixed size at 512 with $L_2$ normalization. For a fair comparison with related methods, we use the same mini-batch sampling strategy with batch size 128. The only difference in experimental setup is that we focus on finetuning from different pretrained models. For finetuning, instead of using SGD/Adam optimizers for a continued loss optimization, we use MetricOpt to finetune $d=128$ conditioning parameters of FiLM layers for only $T=200$ steps. Such finetuning relies on a value function optimized for the evaluation metric of average Recall@k.

For good performance, we use our value function to augment state-of-the-art surrogate losses -- FastAP~\cite{Cakir_2019_CVPR} and Recall loss~\cite{Rolinek_2020_CVPR}. They are derived by histogram binning approximation and ranking function interpolation respectively. Table~\ref{tb:sop} confirms our consistent improvements over both surrogate losses due to direct metric optimization.

\subsection{Object detection}

We adopt the commonly-used Faster R-CNN~\cite{faster_rcnn} framework for object detection. Experiments are conducted on the Pascal VOC dataset~\cite{VOC2010} where the VOC 07 or VOC 07+12 trainval sets are used for training, and VOC 07 test set is used for testing (4952 images with 20 object categories). We use the ResNet-50~\cite{He2016DeepRL} backbone and standard hyperparameters (\eg,~batch size, anchor setting) for Faster R-CNN. Our MetricOpt finetunes $d=128$ conditioning parameters of FiLM layers for $T=200$ steps. The finetuning goal is to improve the $AP^{50}$ metric (average precision for boxes with at least 50\% IoU against groundtruth).

Table~\ref{tb:voc} shows MetricOpt-based finetuning is able to obtain notable gains in $AP^{50}$ over two baselines, using either the standard cross entropy loss or state-of-the-art AP loss~\cite{Rolinek_2020_CVPR}. This validates again the effectiveness of MetricOpt as a consistent metric improver but with low cost (hundreds of finetuning steps). Supplementary materials include an interesting observation that our metric-driven finetuning has non-trivial impact on the two-staged Faster R-CNN detector (\eg,~improves recall of the region proposal module). One future work is a detailed investigation of different impacts from varying metrics.

\begin{table}[t]
\caption{Recall(\%)@k on Stanford Online Products dataset.}
\label{tb:sop}
\begin{center}
\begin{small}
\vskip -0.1in
\begin{tabular}{lcccc}
\hline
k & 1 & 10 & 100 & 1000 \\
\hline
FastAP~\cite{Cakir_2019_CVPR} & 76.4 & 89.1 & 95.4 & 98.5 \\
Recall loss~\cite{Rolinek_2020_CVPR} & 78.6 & 90.5 & 96.0 & 98.7 \\
\hline
MetricOpt + FastAP~\cite{Cakir_2019_CVPR} & 79.6 & 91.2 & 96.3 & 98.9 \\
MetricOpt + Recall loss~\cite{Rolinek_2020_CVPR} & \textbf{80.4} & \textbf{91.8} & \textbf{96.5} & \textbf{99.0} \\ \hline
\end{tabular}
\end{small}
\end{center}
\vskip -0.1in
\end{table}

\begin{table}[t]
\caption{Object detection performance in terms of $AP^{50}$ on Pascal VOC 07 test set. CE stands for Cross-Entropy loss.}
\vskip -0.1in
\label{tb:voc}
\begin{center}
\begin{small}
\vskip -0.1in
\begin{tabular}{c|cccc}
\hline
\multirow{2}{*}{Training}  & \multirow{2}{*}{CE} & \multirow{2}{*}{AP loss~\cite{Rolinek_2020_CVPR}} & MetricOpt & MetricOpt\\
 & & & + CE & + AP loss~\cite{Rolinek_2020_CVPR} \\
\hline
07 & 74.2 & 75.7 & 77.8 & \textbf{78.3} \\
07+12 & 80.4 & 81.4 & 82.3 & \textbf{82.6} \\
\hline
\end{tabular}
\end{small}
\end{center}
\vskip -0.1in
\end{table}


\section{Conclusion}

We present a new method to optimize black-box evaluation metrics using a differentiable value function. The value function is meta-learned to provide useful supervision or gradients about metric to augment any user-specified loss. We show it is easy to apply the value function to existing optimizers for gradient-based optimization, which consistently improves over the given loss. The resulting \mbox{MetricOpt} approach is efficient and achieves state-of-the-art performance in various metrics. MetricOpt also generalizes well across tasks and model architectures.

{\small
\bibliographystyle{ieee_fullname}
\bibliography{ref}
}

\clearpage
\appendix

\setcounter{figure}{0}
\setcounter{table}{0}

\makeatletter 
\renewcommand{\thefigure}{S\@arabic\c@figure}
\renewcommand{\thetable}{S\arabic{table}}
\makeatother

\section*{Supplementary Material}

\begin{figure}[t]
\begin{center}
\centerline{\includegraphics[width=1.0\columnwidth]{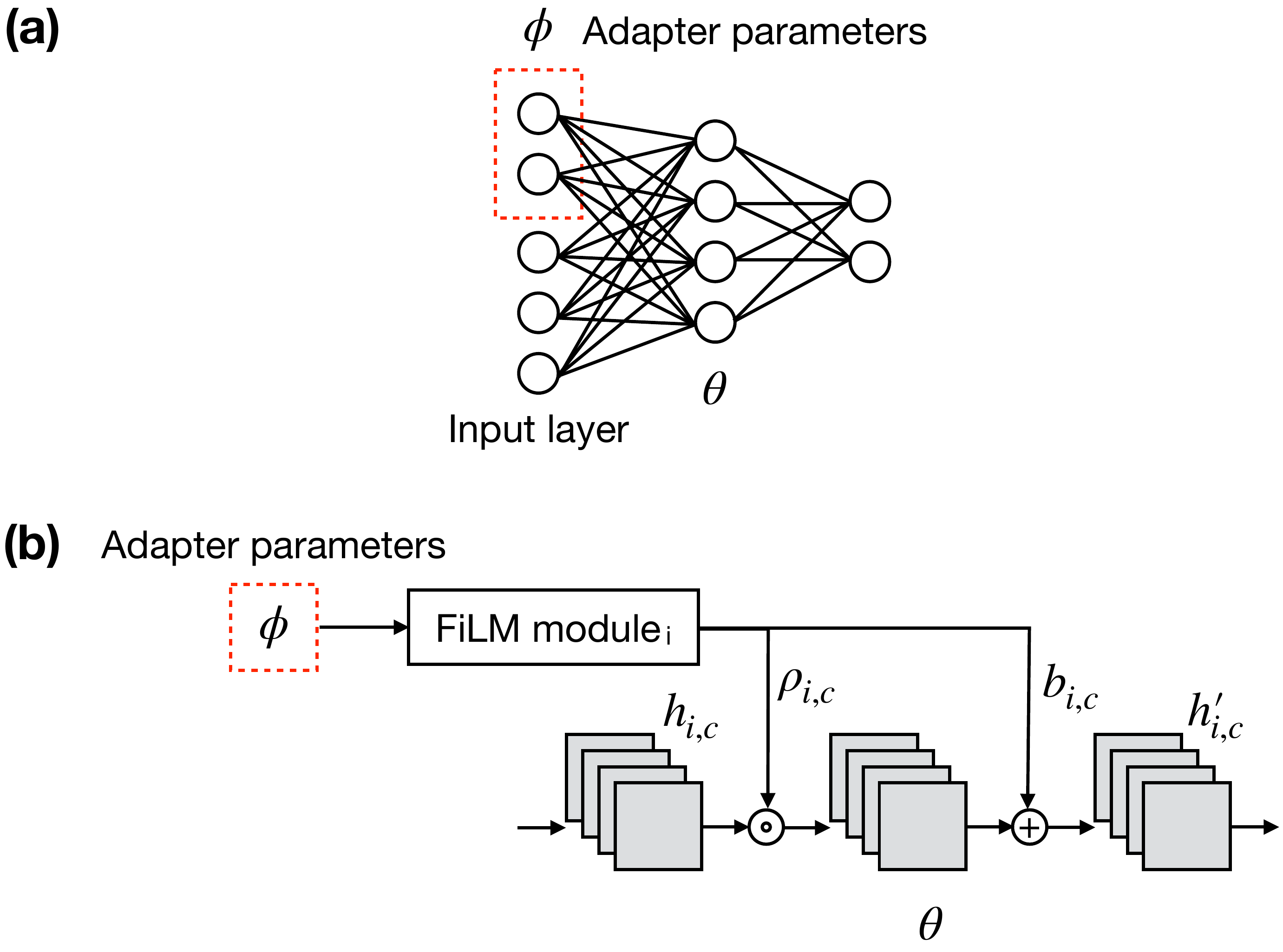}}
\caption{Adapter parameters $\phi$ that modulate the pre-trained network weights $\theta$. (a) For fully connected networks, $\phi$ are the dynamic biases concatenated at the input layer. (b) For convolutional networks, $\phi$ are the input vectors of FiLM layers~\cite{perez2018film} that linearly transform each feature map $h_{i,c}$ with scale $\rho_{i,c}$ and shift $b_{i,c}$ parameters.}
\label{fig:adapter_module}
\end{center}
\end{figure}

\section{Adapter Module Visualization}

For the parameter efficiency of model finetuing, we follow~\cite{zintgrafcavia} to use an adapter module to modulate the pre-trained network weights $\theta$. Specifically, we finetune a small set of parameters $\phi$ of an adapter module. Fig.~\ref{fig:adapter_module} illustrates the adapter modules for both fully connected and convolutional networks. We found the two types of adapter parameters achieve a good performance-efficiency tradeoff, and use them to learn our value function throughout the paper.

\section{MetricOpt with Jointly Learned Optimizer}

In the main paper, we introduce a well-performing \mbox{MetricOpt} that combines our learned value function with hand-crafted optimizers SGD/Adam. Here we show it is also possible to jointly learn the value function \emph{and} the optimizer for better performance. Training signals come from a given surrogate loss and the differentiable value function learned on the fly (as a metric supervision). The main product of such joint training is a neural network parameterized optimizer. During meta-testing, only the learned optimizer is used for a new finetuning task, while the learned value function is dropped. Algorithm S1 shows the pseudocode of a fully learned MetricOpt.


\begin{figure}[!t]
\begin{center}
\centerline{\includegraphics[width=0.95\columnwidth]{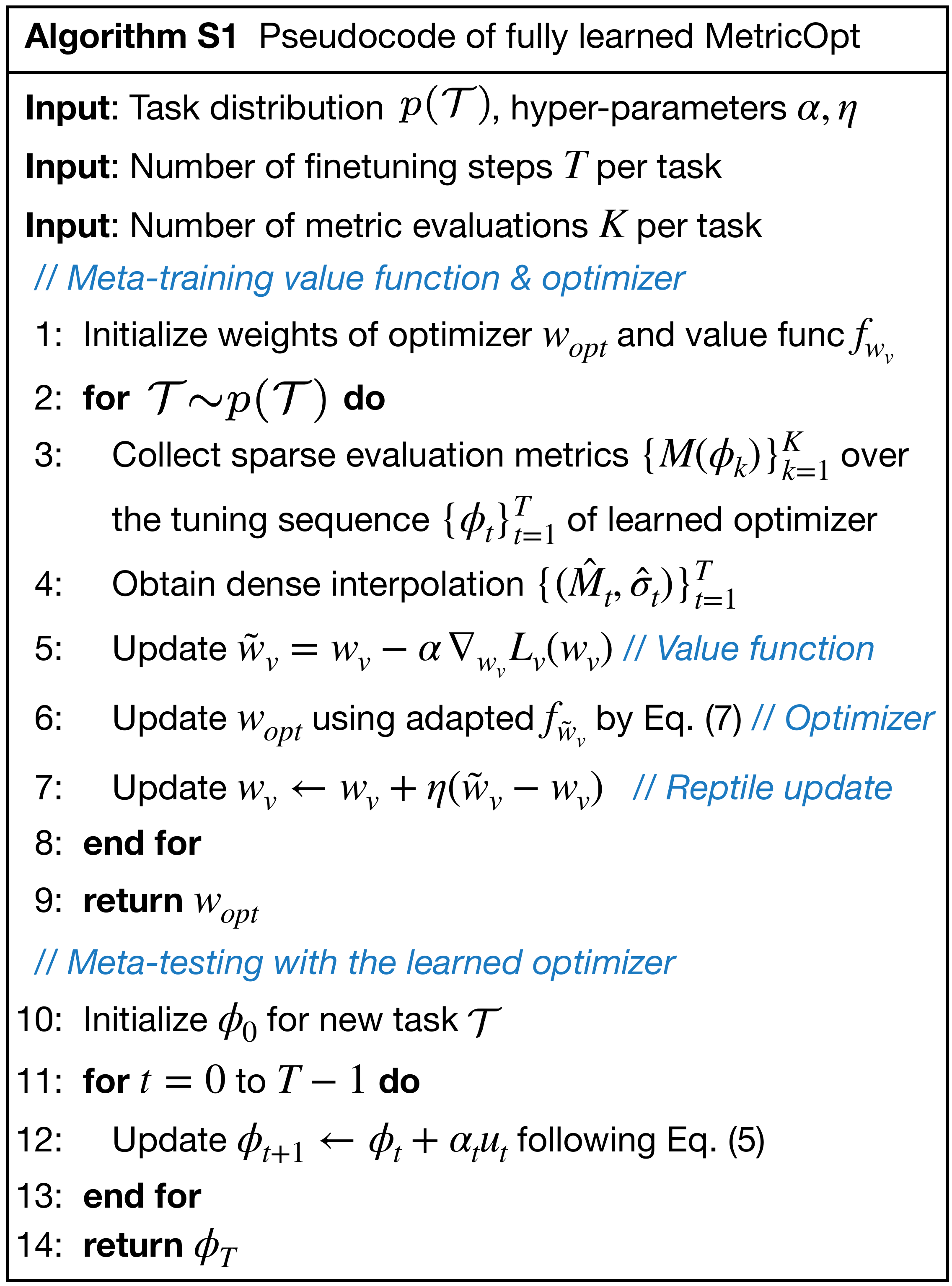}}
\label{fig:algorithm_supplement}
\end{center}
\end{figure}

The current go-to method for optimizer parameterization is to leverage an LSTM network~\cite{NIPS2016_6461,Wichrowska2017a}. Unfortunately, LSTM training suffers from either biased or exploding gradients during truncated backpropagation through unrolled optimization~\cite{metz19a}. Here we use a simple MLP with weights $w_{opt}$ as in~\cite{metz19a}. The MLP-based optimizer $m_{w_{opt}}$ can produce the update direction $u_t$ and learning rate $\alpha_t$ to recurrently update our adapter parameters $\phi_t$:
\begin{equation}  
\begin{split}
\phi_{t+1} &= \phi_t + \alpha_t u_t, \\
[\alpha_t,u_t] &= m_{w_{opt}}(\nabla_t, \bar{\nabla}_t, \phi_t, \ell_t, \Delta \ell_t, \mathcal{M}_t, \Delta \mathcal{M}_t), \\
\end{split}
\label{eqs1}
\end{equation}
where optimizer inputs are the loss gradient $\nabla_t=\nabla_{\phi} \ell(\phi_t)$, exponential running average $\bar{\nabla}_t$, current parameters $\phi_t$, scale-normalized loss $\ell_{t}$ and metric $\mathcal{M}_{t}$ and their relative changes $\Delta \ell_t$ and $\Delta \mathcal{M}_t$ from the respective moving averages.

As mentioned above, we train $w_{opt}$ with supervisions of both metric $L_{metric}$ and loss $L_{loss}$:
\begin{eqnarray}  
\!\!\!\!\!\!\!\!\!\!\!\!\! && L_{opt}(w_{opt})  = \lambda L_{metric}(w_{opt}) + L_{loss}(w_{opt}), \\
\!\!\!\!\!\!\!\!\!\!\!\!\! && L_{metric}  =\frac{1}{T} \sum_{t=1}^T \log\left( 1+\exp \left( \frac{\beta(\hat{\mathcal{M}}_t-\hat{\mathcal{M}}_{t'})}{\hat{\mathcal{M}}_t} \right) \right), \nonumber \\
\!\!\!\!\!\!\!\!\!\!\!\!\! && \hat{\mathcal{M}}_t = f_{w_v}(\phi_t), \; t'=\underset{i<t}{\arg\min} \hat{\mathcal{M}}_i, \nonumber \\
\!\!\!\!\!\!\!\!\!\!\!\!\! && L_{loss}  = \frac{1}{T} \sum_{t=1}^T \left( \log(\ell(\phi_t)+\epsilon) - \log(\ell(\phi_0)+\epsilon) \right), \nonumber
\label{eqs2}
\end{eqnarray}
where hyper-parameters $\lambda=50$ and $\beta=T/2$, and the small positive constant $\epsilon$ is introduced to avoid numerical instability. Note $\hat{\mathcal{M}}_t$ is the metric prediction from our value function $f_{w_v}$. The $L_{metric}$ term encourages relative metric improvements for the whole optimization sequence in a logarithmic form. On the other hand, the $L_{loss}$ term evaluates the average log loss (offset by the initial value). This term encourages a low loss at convergence while still providing loss training signals at every step.

One benefit of the multi-task objective $L_{opt}$ is that it continues to penalize metric deterioration when minimizing the loss at convergence. However, the objective surface of $L_{opt}$ can be extremely non-smooth~\cite{metz19a}, and we are likely to obtain noisy derivatives through the unrolled optimization process. Here, we follow~\cite{metz19a} to train our optimizer $m_{w_{opt}}$ using a variational loss as a smoothed objective:
\begin{equation}  
\mathbb{E}_{\tilde{w}_{opt} \sim \mathcal{N}(w_{opt},\varepsilon^2 I) } \; L_{opt}(\tilde{w}_{opt}),
\label{eqs3}
\end{equation}
where $\varepsilon^2=0.01$ is a fixed variance. This smoothing helps stabilize optimizer training. Actual training is based on the two unbiased gradient estimators in~\cite{metz19a}, with the same learning rate setting.

\textbf{Computational complexity} Our main paper confirms the empirical advantages of jointly learning the value function and optimizer. The resulting \textbf{MetricOpt (learned)} outperforms our default \textbf{MetricOpt (SGD)} approach when optimizing the image classification metrics MCR and AUCPR on CIFAR-10 dataset. But it is worth noting that such performance gains come at a larger meta-training cost. Specifically, joint training needs more meta-iterations to converge than meta-training the value function alone in our default approach ($\sim$5k vs. $<$2k, with often more than 3$\times$ longer training time). The joint training time is dominated by that of optimizer learning (with costly unrolled derivatives). On the side of value function learning, the corresponding cost is much smaller. During meta-testing, the optimization cost will not affected by value function at all since only the learned optimizer is used.

\textbf{More results} We further benchmark the fully learned \mbox{MetricOpt} on the large-scale ImageNet dataset. Table~\ref{tb:imagenet_learned_optimizer} validates the advantage of MetricOpt (learned) over MetricOpt (SGD) again. The comparison also confirms the efficacy of our value function for gradient-based metric optimization.

\begin{table}[t]
\caption{Top-1 and Top-5 classification accuracies (\%) on ImageNet. All methods use the same NASNet-A network.}
\label{tb:imagenet_learned_optimizer}
\begin{center}
\begin{small}
\begin{tabular}{lcc}
\hline
Method & Top-1 & Top-5 \\
\hline
RMSProp + cross-entropy & 73.5 & 91.5 \\ 
MetricOpt (SGD) & 74.9 & 92.7 \\
MetricOpt (learned) & \textbf{75.0} & \textbf{93.0} \\ \hline
\end{tabular}
\end{small}
\end{center}
\end{table}

\begin{figure}[t]
\begin{center}
\centerline{\includegraphics[width=1.0\columnwidth]{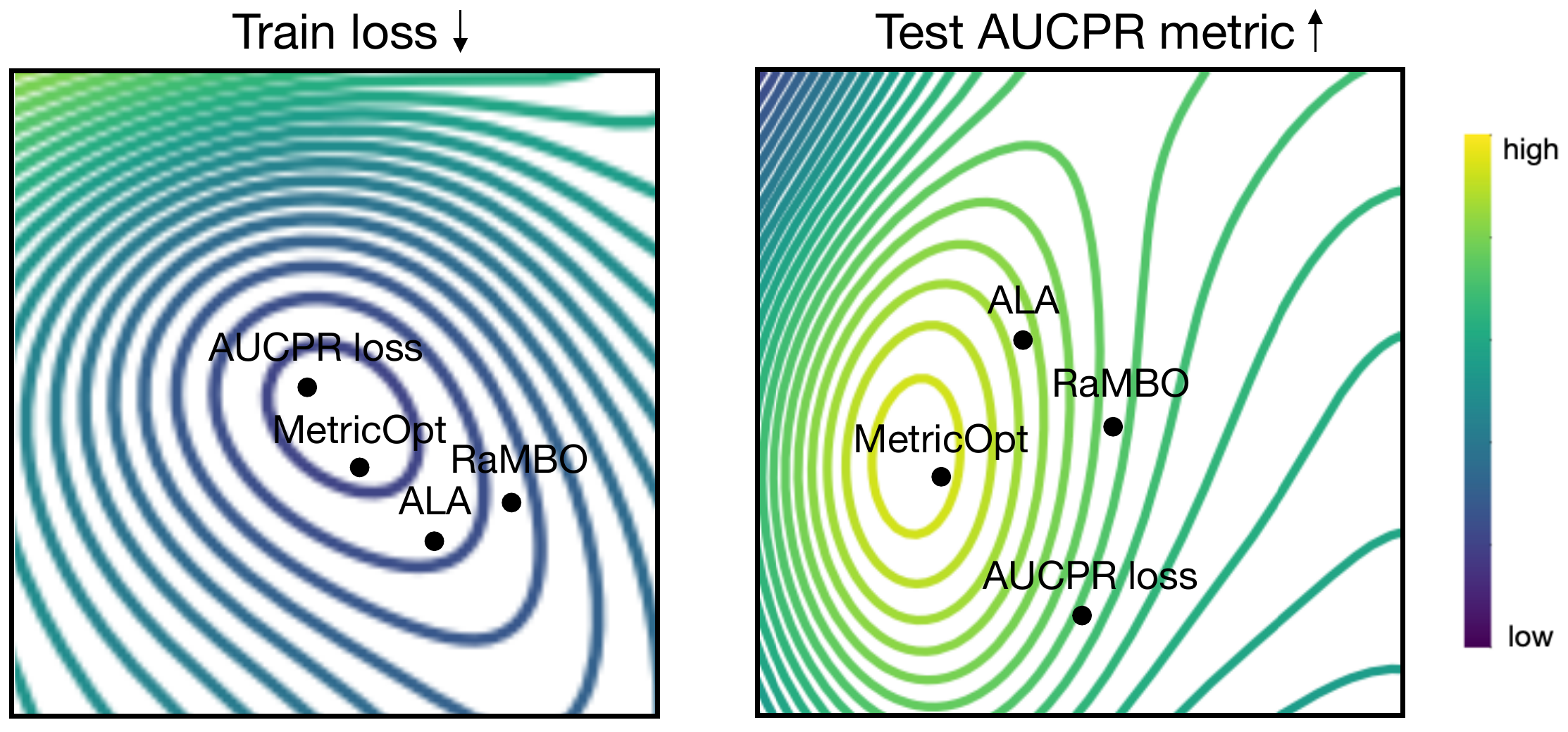}}
\caption{Train loss and test metric surfaces (visualized by the toolbox in~\cite{visualloss}) for optimizing the AUCPR metric on CIFAR-10. We compare the solutions obtained from the hand-designed AUCPR loss~\cite{pmlr-v54-eban17a}, Adaptive Loss Alignment (ALA)~\cite{ALA_2019}, RaMBO (with gradient interpolation)~\cite{Rolinek_2020_CVPR} and our MetricOpt.}
\label{fig:compare_landscape}
\end{center}
\vskip -0.4in
\end{figure}

\section{Understanding Differences among Recent Methods for Metric Optimization}

In the field of non-differentiable metric optimization, there are recent strong methods that have been compared against in the main paper. Here we provide more analyses and visualizations to explain the advantages of our \mbox{MetricOpt} method. Recall MetricOpt learns a differentiable value function that generates reliable gradient estimates of metric to augment loss gradients. How does this impact the training dynamics on the optimization landscape? Fig.~\ref{fig:compare_landscape} gives some hints by visualizing the typical optimization surfaces of train loss and test metric, as well as the converged solutions proposed by different methods. The compared methods are of three types: loss function~\cite{pmlr-v54-eban17a} designed to approximate the target metric, adaptively learned loss ALA~\cite{ALA_2019}, and RaMBO (with gradient interpolation)~\cite{Rolinek_2020_CVPR}.

\begin{figure*}[!t]
\begin{center}
\centerline{\includegraphics[width=1.0\linewidth]{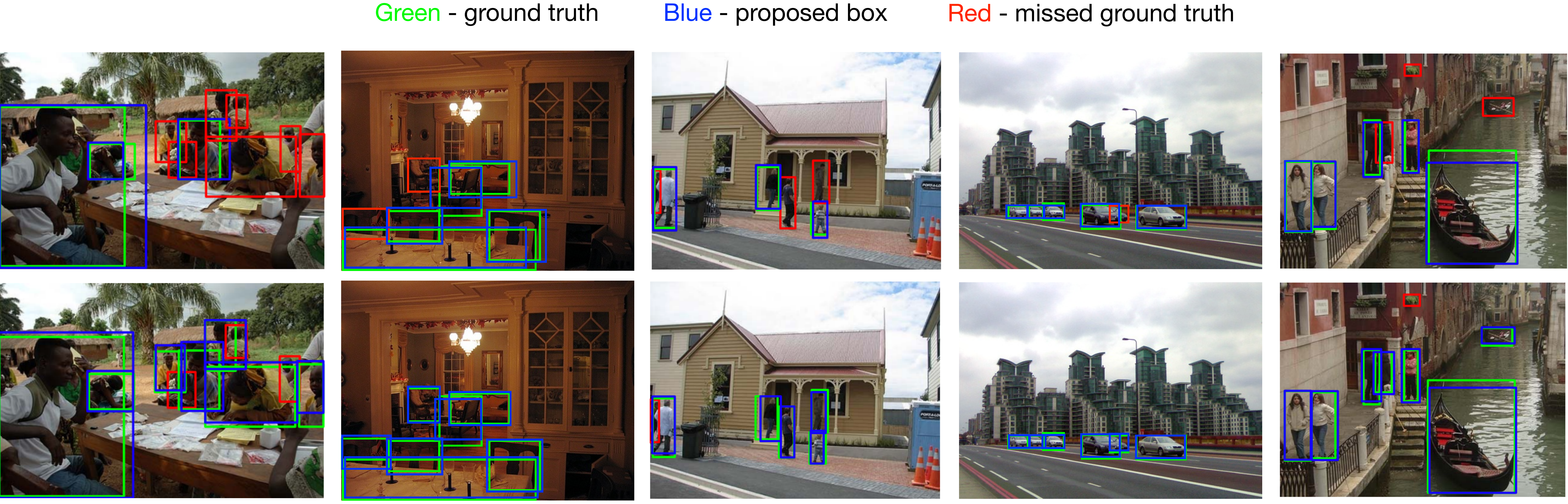}}
\vskip 0.1in
\caption{Visual comparison between the region proposal networks of two Faster R-CNN baselines. Top row: object proposals (after IoU thresholding) with standard Cross-Entropy loss (CE). Bottom row: object proposals with MetricOpt + CE. By direct optimization of the target metric $AP^{50}$, MetricOpt is found to improve the region proposal module of the two-staged Faster R-CNN detector.}
\label{fig:detection_supplement}
\end{center}
\vskip -0.25in
\end{figure*}

We observe from the figure that:
\begin{itemize}[leftmargin=10pt]
\setlength{\itemsep}{0pt}
\setlength{\parsep}{0pt}
\setlength{\parskip}{0pt}
\item loss and metric surfaces are different, verifying the need for some form of metric supervision during optimization.
\item All compared methods achieve low training loss values, but their testing metrics have notable differences. This confirms the observation in~\cite{draxler18a} that low-loss solutions form a connected manifold, and further adds that they tend to be distinct in the metric space due to different ways of metric approximation. 
\item Specifically, both human-designed and learned losses (\ie,~AUCPR and ALA losses) need a relaxed surrogate space to approximate metrics. Obviously, performance will highly depend on the quality of such surrogate relaxations. For RaMBO, gradient interpolation is conducted via black-box differentiation which can be critically affected by the sparsity of supervision signals and interpolation details. Our MetricOpt sidesteps these challenges by a direct function approximation of target metric (with value function). The resulting metric supervision can adjust the optimization trajectory towards better metric, leading to a solution off trajectories purely based on surrogate losses. Fig.~\ref{fig:compare_landscape} shows that \mbox{MetricOpt} is able to converge to the best performance metric while still maintaining a low loss.
\end{itemize}

\section{More Ablation Studies}

\begin{figure}[t]
\begin{center}
\centerline{\includegraphics[width=1.0\columnwidth]{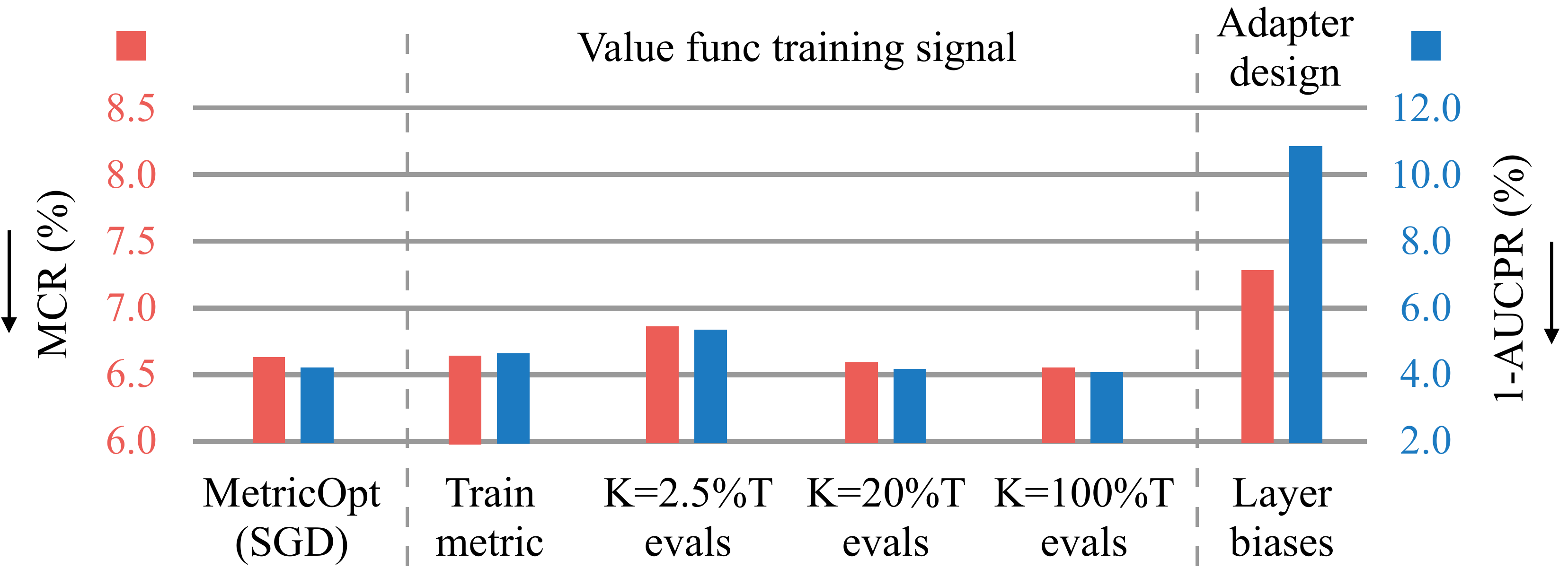}}
\vskip 0.1in
\caption{Ablation studies of value function for optimizing Miss-Classification Rate (MCR) and Area Under the Precision Recall Curve (AUCPR) on CIFAR-10 dataset. The AUCPR metric undergoes a $1-x$ conversion (thus lower is better). In terms of both metrics, we compare different training signals and input parameters for the value function.}
\label{fig:ablation_supplement}
\end{center}
\vskip -0.4in
\end{figure}

Fig.~\ref{fig:ablation_supplement} ablates the different training aspects of our value function. We observe that:
\begin{itemize}[leftmargin=10pt]
\setlength{\itemsep}{0pt}
\setlength{\parsep}{0pt}
\setlength{\parskip}{0pt}
\item Learning with evaluation metrics from a subset of training data with the same size of validation set (default) barely hurts performance. This suggests our advantages mainly come from direct metric optimization, not just from learning validation statistics.
\item By default, we collect $K=5\%T$ evaluation metrics from each finetuning task with $T$ iterations and then interpolate sparse metrics temporally for value function learning. We found a smaller $K$ (\eg,~$K=2.5\%T$) hinders effective value function learning (hurts final performance too). On the other hand, larger $K$s slightly improve performance, but lead to increased cost for dense metric evaluation. Note when $K=100\%T$, we collect evaluation metrics from all iterations, without using any metric interpolation. This proves as unnecessary since the resulting gains are pretty marginal.
\item Regarding the value function input, we choose to use the compact adapter parameters $\phi$ by default (with size 128 in the case here). Other options exist, including using a small portion of the main network $\theta$ like the biases of last layer. Results indicate last layer biases are not enough to model metrics well, while using multi-layer biases becomes more parameter-inefficient with even worse results. In our early experiments, we failed to learn value function from the entire last layer for the same reason. But when learning with $\phi$, performance is reasonably robust to $\phi$'s size which does not need careful tuning.
\end{itemize}

\section{More Object Detection Results}

The main paper (Table 6) shows notable improvements over the Faster R-CNN detector by our direct metric optimization during finetuning. But what contributes to the gains, and how does finetuning impact the different modules of Faster R-CNN (\ie,~region proposal network and detection network)? Fig.~\ref{fig:detection_supplement} sheds some light on these questions. Through finetuning for the AP metric, we observe the region proposal network seems to have an improved recall on the proposed bounding boxes. Intuitively, such improved object proposals should benefit the following detection module. This inspires an interesting future work, which is to investigate different impacts on a multi-staged pipeline when optimizing for different metrics.

\end{document}